%% file: mvp.tex
\documentclass[letterpaper, 10 pt, conference]{ieeeconf}  

\IEEEoverridecommandlockouts                              

\let\labelindent\relax

\usepackage{amsmath}
\usepackage{amsfonts}
\usepackage{listings}
\usepackage{courier}
\usepackage{booktabs}
\usepackage{textcomp}
\usepackage{color,soul}
\usepackage{subcaption}
\usepackage{caption}
\usepackage{algorithm}
\usepackage{algorithmic}
\usepackage{tabularx}
\usepackage{caption}
\usepackage{siunitx}
\usepackage{enumitem}
\usepackage[table]{xcolor}
\usepackage{multirow}
\usepackage{booktabs}
\usepackage{makecell}
\usepackage{mwe}
\usepackage{listings}
\usepackage{adjustbox}
\usepackage{breakcites}
\usepackage{siunitx}
\usepackage{verbatim}
\usepackage{placeins}
\usepackage{mathrsfs}
\usepackage{float}
\usepackage{hyperref}

\title{\LARGE \bf
Multiple View Performers for Shape Completion
}

\author{David Watkins-Valls$^{1}$, Peter Allen$^{1}$, Krzysztof Choromanski$^{2}$, Jacob Varley$^{2}$, and Nicholas Waytowich$^{3}$
\thanks{$^{1}$Department of Computer Science, Columbia University, New York, NY, USA,  {\tt\small\{davidwatkins,allen\}@cs.columbia.edu}}%
\thanks{$^2$Robotics at Google. {\tt\small\{kchoro,jakevarley\}@google.com}}%
\thanks{$^3$U.S. Army Research Laboratory, Baltimore, MD, USA. {\tt\small nicholas.r.waytowich.civ@mail.mil}}%
\thanks{This research was sponsored by the Army Research Laboratory and was accomplished under Cooperative Agreement Number W911NF-18-2-0244. The views and conclusions contained in this document are those of the authors and should not be interpreted as representing the official policies, either expressed or implied, of the Army Research Laboratory or the U.S. Government. The U.S. Government is authorized to reproduce and distribute reprints for Government purposes notwithstanding any copyright notation herein.}
}

\begin{document}

\maketitle
\thispagestyle{empty}
\pagestyle{empty}

\begin{abstract}
We propose the \textit{Multiple View Performer} (MVP) - a new architecture for 3D shape completion from a series of temporally sequential views. MVP accomplishes this task by using linear-attention Transformers called \textit{Performers} \cite{performer}. Our model allows the current observation of the scene to attend to the previous ones for more accurate infilling. The history of past observations is compressed via the compact associative memory approximating modern continuous Hopfield memory, but crucially of size independent from the history length. We compare our model with several baselines for shape completion over time, demonstrating the generalization gains that MVP provides. To the best of our knowledge, MVP is the first multiple view voxel reconstruction method that does not require registration of multiple depth views and the first causal Transformer based model for 3D shape completion.
\end{abstract}

\input{sections/introduction}
\input{sections/related_work}
\input{sections/mva_model}

\input{sections/experiments}
\input{sections/real_world}
\input{sections/conclusion}

\bibliographystyle{IEEEtran}
\bibliography{IEEEabrv,mvp}

\end{document}

%% file: sections/introduction.tex
\section{Introduction}
Shape completion from a single image or two images is a difficult and important problem (see: \cite{shape-completion-base, shaperetsurvey, moons, tomo, medicalapp, haefner2019photometric, varley2017shapecompletion_iros, Yang18}). Providing more accurate reconstructions of objects helps enable a variety of robotic tasks such as manipulation, collision checking, sorting, and cataloging. Synthesizing multiple images for accurate 3D-reconstruction of an object is even more challenging.

\begin{figure}[!htbp]
    \centering {
        \includegraphics[width=0.95\linewidth]{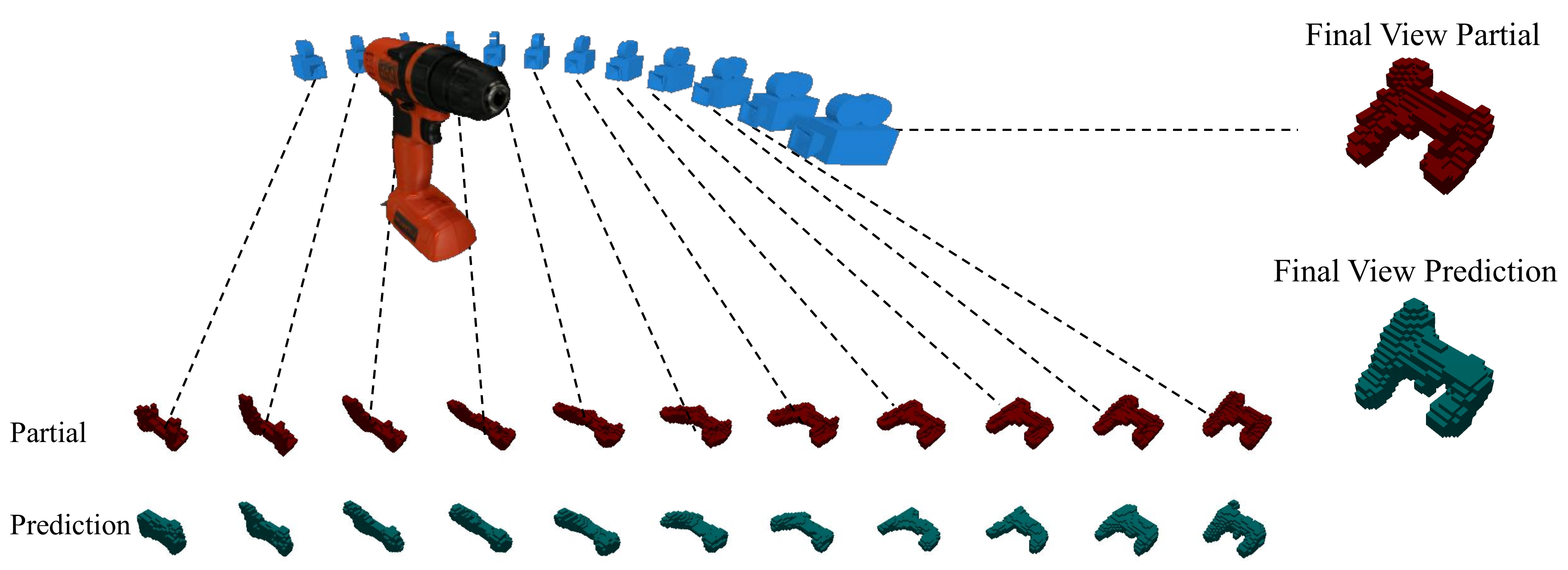}
    }
    \caption{\small{Shown in red are partial views of the target object and in green are prediction of each incremental view using our Performer-based MVP to shape completion. The final prediction, in the top right, is the culmination of multiple successive views contributing to an overall completion of the target object, in this case a drill from the YCB object dataset~\cite{calli2015ycb}.}} \label{fig:multiviewcoverfigure}
\end{figure}

\begin{figure}[!htbp]
    \centering {
        \includegraphics[width=0.95\linewidth]{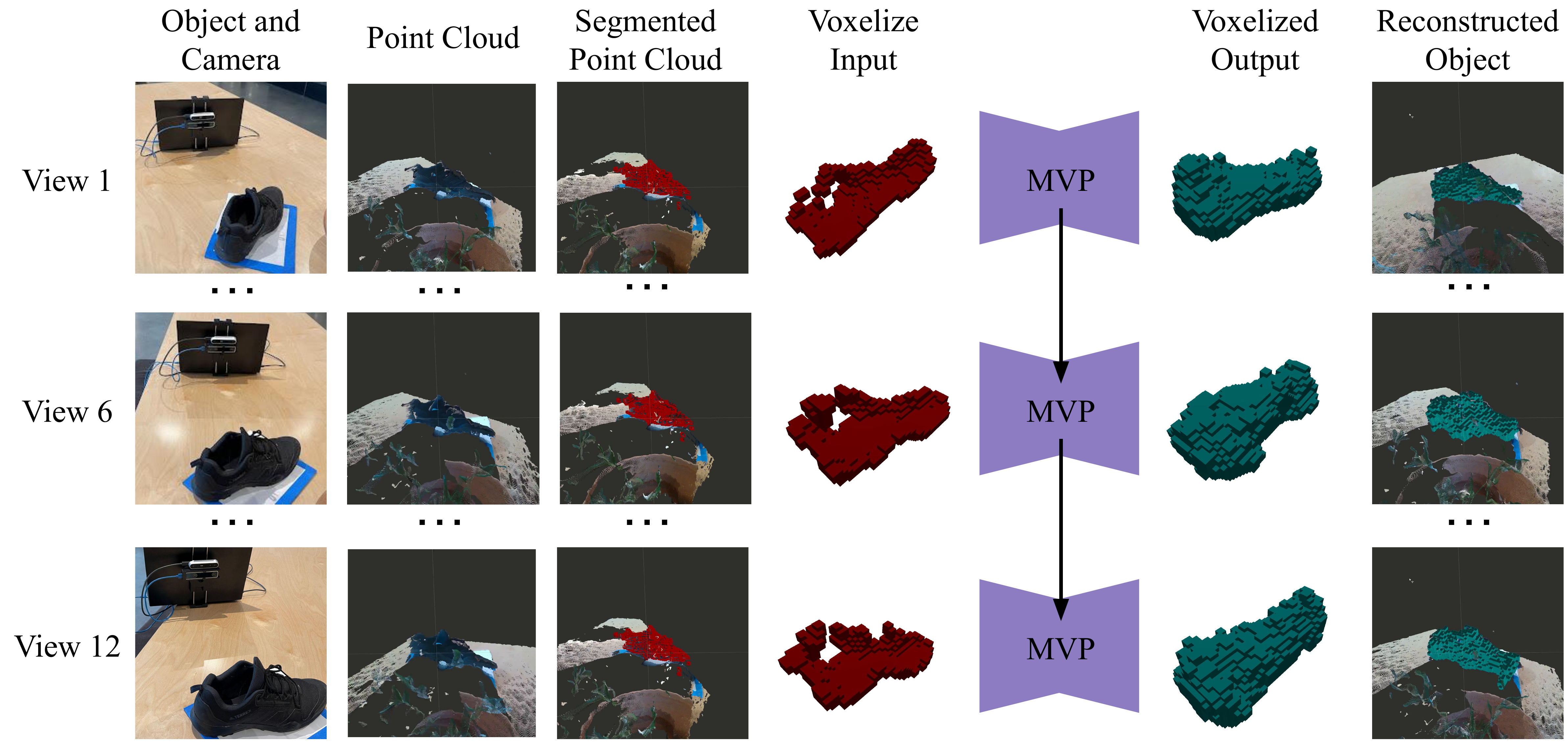}
    }
    \caption{\small{A shoe placed on a table in front of an Intel Realsense d415 camera and a Intel Realsense T265 tracking camera. The T265 camera helps segment the shoe from the environment. Each incremental view of the shoe, shown in red, helps refine the reconstruction of the object, shown in green. This model is trained in simulation and can be used to complete objects in the real world. }} \label{fig:multiviewdemo}
    \vspace{-3mm}
\end{figure}

We propose a novel approach to 3D shape reconstruction, called \textit{Multiple View Performer} (or: MVP) that can be used to complete objects with only two views, or up to an arbitrary number of views, leveraging recently introduced class of scalable linear-attention Transformers \cite{vaswani2017attention}, called \textit{Performers}~\cite{performer,slim-performers}. 
At MVP runtime, 2.5D views about the object or simple scene are captured in a panning motion to create a sweeping snapshot of the object's geometry (see~\autoref{fig:multiviewcoverfigure}), or from a still camera in the case of moving objects. For each of these views, the causal performer block updates its corresponding compact associative memory (approximating modern continuous Hopfield memory~\cite{hopfield}), effectively improving MVP's understanding of an object and consequently - the overall shape estimation. Crucially, the size of  the aforementioned compact associative memory is independent from the number of views it consumed (see: Section~\ref{sec:memory} for more details).
When a completion is requested, the current observation implicitly interacts with all the previous observations through that compact memory for its more accurate infilling.
Due to the Performer block's ability to memorize multiple views, it can also remember objects that are no longer visible or utilize newly revealed views of objects that were previously hidden. Through our results, we will show that the proposed MVP system is able to generalize better both for single view and multiple view reconstruction without requiring the registration of multiple views of the object. We will show that this shape completion system is able to perform better or on par versus an LSTM-based system and an attention-based system. This system can be used for many different robotics tasks, such as grasping, stacking, and collision avoidance.  We show a real world demonstration of how this could be used with a camera fixed to a robot in~\autoref{fig:multiviewdemo}. We also demonstrate, using a simulated BarrettHand, that this shape completion system can be used for grasp planning. 

%% file: sections/related_work.tex
\section{Related Work}

A single-depth sensor can be moved to capture multiple views of an object, but aligning those views remains hard. Work done by Tremblay et al.~\cite{dope} has shown that a grasp planner can plan and execute a grasping task given a mesh and an estimated pose of that mesh. Other work from Varley et al.~\cite{varley2017shapecompletion_iros} and Watkins-Valls et al.~\cite{watkinsvalls2022mobile} have shown that a single view or two views of an object can be used to reconstruct the object for grasp planning via GraspIt!~\cite{miller2004graspit}. Other work by Wang et al.~\cite{Wang_2019_CVPR} has shown that objects can be manipulated via dense fusion of multiple images of the object by requiring registration between those images.  

There has been much successful research in utilizing continuous streams of visual information like Kinect Fusion~\cite{newcombe2011kinectfusion} or SLAM~\cite{thrun2008simultaneous} to improve models of 3D objects for manipulation, an example being~\cite{krainin2010manipulator,krainin2011autonomous}. A popular approach (see for instance RTABMap~\cite{labbe2019rtab}) is to perform visual RGBD slam to register multiple point clouds. It can be challenging for a robotic agent to denoise registered images and eliminate potentially introduced errors into a machine-learned CNN. This type of noise is difficult to model in simulation. A 3D-CNN~\cite{3dcnns} can be used to enable robust shape estimation by leveraging multiple unregistered view, meaning that each image of an object is kept in its respective image frame. 

Zha et al.~\cite{semantic3dreconstruction} developed an approach to build 3D models of unknown objects based on a depth camera observing the robot's hand while moving an object. This strategy integrates both shape and appearance information into an articulated ICP approach to track the robot's manipulator and the object while improving its 3D model. Similarly, another work~\cite{hermann2016eye} attaches a depth sensor to a robotic hand and plans grasps directly in the sensed voxel grid. These methods improve the 3D-models of the objects using only a single sensory modality but from multiple points in time. ]

Modern work in shape reconstruction has centered around pose estimation of known objects~\cite{qi2017pointnet,qi2017pointnet++}. The corresponding techniques utilize multi-layer-perceptrons (MLP) to map points into dense representations and perform classification tasks. They can be useful for scene segmentation, part identification, or pose estimation. Other work has divided the reconstruction and pose estimation into two separate steps to allow for a more refined mesh prediction of a set of objects on a table~\cite{irshad2022centersnap}. 


New work in rendering unseen images using implicit representations is catalyzing new research in the field. Sitzmann et al. \cite{siren} analyzed the effectiveness of generating meshes, images, and even sound from input data via implicit representations and novel use of sine as an activation function. Other approaches include: generating novel views of an environment from a single view~\cite{eslami2018neural} and utilizing text-based mesh-generation applying implicit representations~\cite{texttomesh2022khalid}. 

Attention-based architectures for 3D shape reconstruction is an increasingly growing area of research
\cite{salvi, boyang, mvt,danwang}, yet most of the works focus on: (1) the bidirectional (encoder) setting to determine which regions of the input to focus on, (2) the reconstruction from the 2D images, or (3) are unable to realign the reconstruction into the real world. The attractive memorization aspect of the unidirectional (causal) attention for 3D point clouds is not well explored. Memory-based models were so far exploited mainly in the context of RNNs \cite{choy}. Multiple view networks such as~\cite{9022145,dai2017shape,choy20163d,Peng_2022_CVPR} are designed to reconstruct an object in a known image frame rather than completing the object in the image frame it appears in. 

%% file: sections/mva_model.tex
\section{The Multiple View Performer (MVP) Architecture}
\vspace{-3mm}
\begin{figure}[h]
    \centering
    \begin{subfigure}{\linewidth}
    \includegraphics[width=\linewidth]{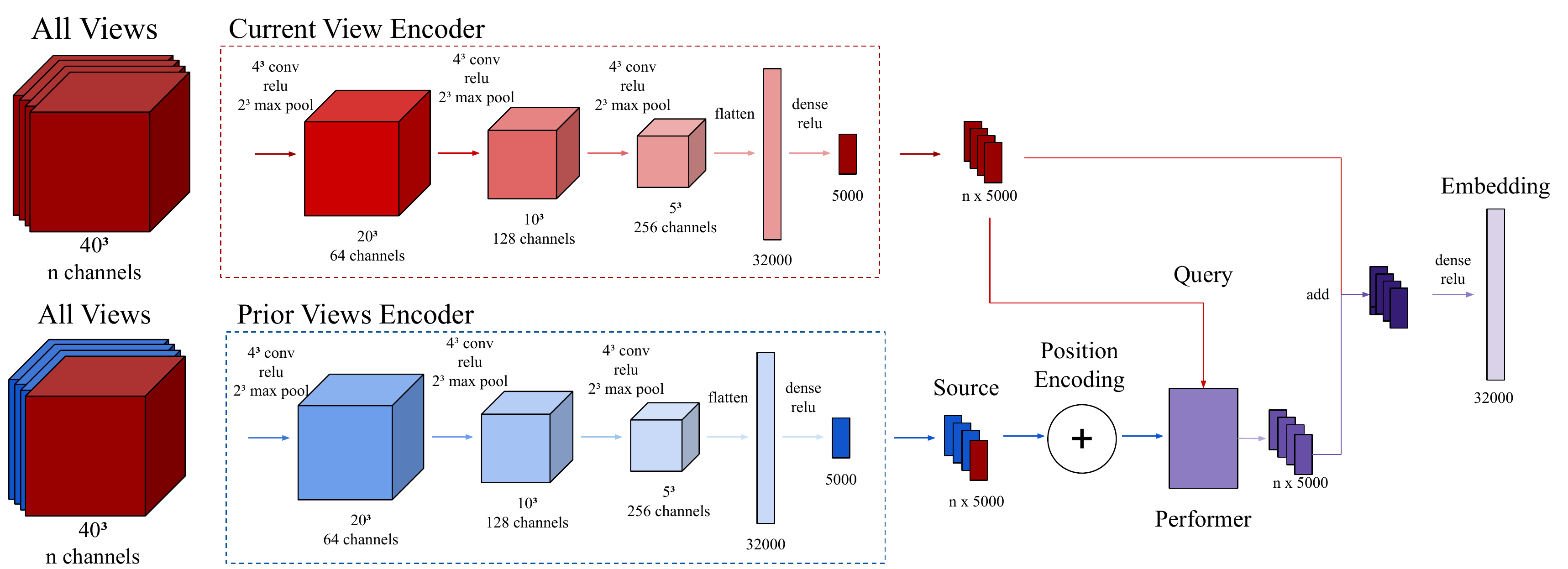}
    \caption{MVP Encoder architecture.}
    \label{fig:performer_encoder_v1}
    \end{subfigure}
    
    \begin{subfigure}{\linewidth}
    \includegraphics[width=\linewidth]{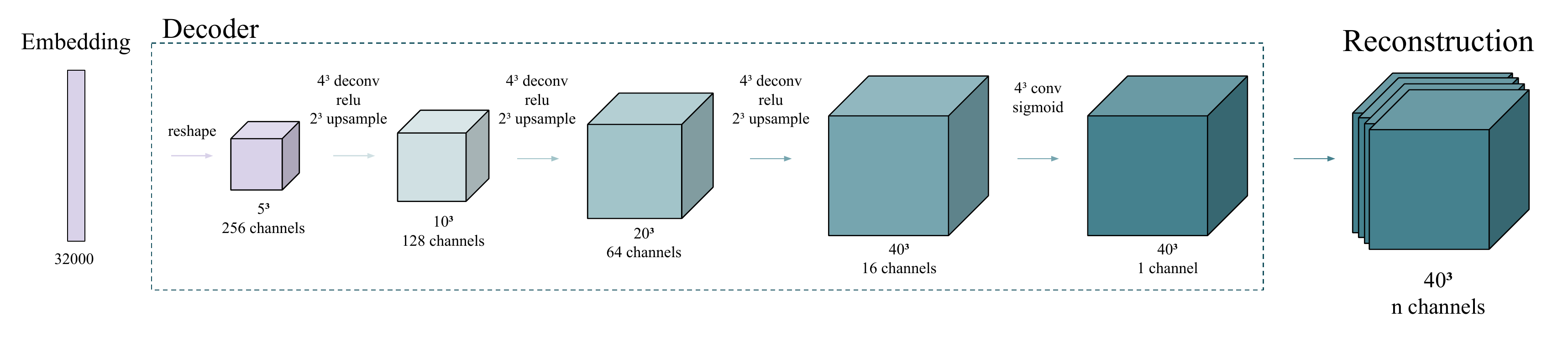}
    \caption{MVP Decoder architecture.}
    \label{fig:performer_decoder_v1}
    \end{subfigure}

    \caption{\small{The MVP-network takes multiple unregistered views of an object to produce a reconstruction. (a) Its encoder-part has a two-tower structure with one tower acting solely on the input frame and the other leveraging the context given by Performer's associative memory and encapsulating all frames seen so far. As opposed to a regular Transformer, the size of this memory is independent from the number of frames seen so far. The second tower enables the input frame to attend to the compact summarization of the previous frames. (b) The outputs of both encoders are added and fed to the decoding layer.}} \label{fig:performerv5architecture}
\end{figure} 

To leverage multiple views, an architecture that can operate on the ordered sequence of embeddings is required. There are many natural candidates for such an architecture, including LSTMs \cite{lstm-1, lstm-2}, GRUs \cite{gru, gru-1}, and attention-based models. MVP utilizes the last class, given unprecedented success of Transformers \cite{vaswani2017attention} in modeling sequential data. It uses two encoders: (1) the first one acts solely on the new input frame, (2) the second one feeds causal Performer with that frame (enriched with its positional encoding), effectively enabling it to attend to all previous frames through Performer's compact associative memory. The resulting two embeddings are then added to each other and this aggregate embedding is fed to the decoder to reconstruct the object geometry (see: Fig. \ref{fig:performerv5architecture}). 

Using Performers instead of regular Transformers is a pragmatic choice.  The latter ones are characterized by linear in the number of the observed frames memory size as well as memory update (per new frame). Instead, Performers guarantee constant in the frame-history length update of the encoder towers and memory size. We explain this memory model in detail in the next section.

\subsection{MVP memory with causal Performers}
\label{sec:memory}

All the vectors in this section are by default row-vectors. We consider a sequence of observations (image frames) $(o_{1},...,o_{L}) \in \mathbb{R}^{40 \times 40 \times 40 \times 1}$, each represented as a voxel grid with occupancy scores. In the attention approach to frame-sequence modeling, each observation $o_{i}$ is associated with a latent representation denoted as $\mathbf{v}_{i} \in \mathbb{R}^{d}$ (called a \textit{value vector}). Furthermore, the \textit{attention} of the observation $o_{i}$ to the observation $o_{j}$ is defined as: $\mathrm{att}_{i,j}=\mathrm{K}(\mathbf{q}_{i},\mathbf{k}_{j})$ for two (learnable) latent encodings corresponding to $o_{i}$ and $o_{j}$, called \textit{query} ($\mathbf{q}_{i} \in \mathbb{R}^{d_{\mathrm{QK}}}$) and \textit{key} $(\mathbf{k}_{j} \in \mathbb{R}^{d_{\mathrm{QK}}})$  respectively and  some fixed kernel $\mathrm{K}:\mathbb{R}^{d_{\mathrm{QK}}} \times \mathbb{R}^{d_{QK}} \rightarrow \mathbb{R}$.
Attention models usually apply \textit{softmax-kernel}, defined as:
$\mathrm{K}_{\mathrm{sfm}}(\mathbf{q}_{i},\mathbf{k}_{j})=\exp(\mathbf{q}_{i}\mathbf{k}_{j}^{\top})$.
Sequence $(o_{1},...,o_{L})$ defines the \textit{memory} $\mathcal{M}$ of the system. 

For the newly coming frame $o_{i}$, the latent representation of the most relevant frame from the memory is retrieved approximately as a convex sum of value vectors for the frames seen so far with the renormalized attention coefficients, i.e. has the following form: 
\begin{equation}
\label{mem-equation}
\mathbf{x}_{i}=\sum_{j=1}^{i} \frac{\mathrm{K}(\mathbf{q}_{i},\mathbf{k}_{j})}{\sum_{l=1}^{i}\mathrm{K}(\mathbf{q}_{i},\mathbf{k}_{l})} \mathbf{v}_{j}.
\end{equation}
\noindent 
This retrieval process can be thought of as a one gradient step (with learning rate $\eta=1$) of the \textit{continuous Hopfield network} with the exponential energy function~\cite{hopfield}. If the keys of the observations are spread well enough, the procedure within a couple of gradient steps converges to the value vector corresponding to the nearest-neighbor of $o$ from $\mathcal{M}$ (with respect to the dot-product similarity in the query-key space). This property is true even for the exponential-size memories.

This approach has a critical caveat though - the memory needs to be explicitly stored. It becomes problematic if a substantial number of observations $L$ are collected since $\mathcal{M}$ grows linearly in $L$ and latent embedding computation from Eq. \ref{mem-equation} clearly takes time linear in $L$. 

\begin{figure}[t]
\vspace{2mm}
    \centering{
        \includegraphics[width=\linewidth]{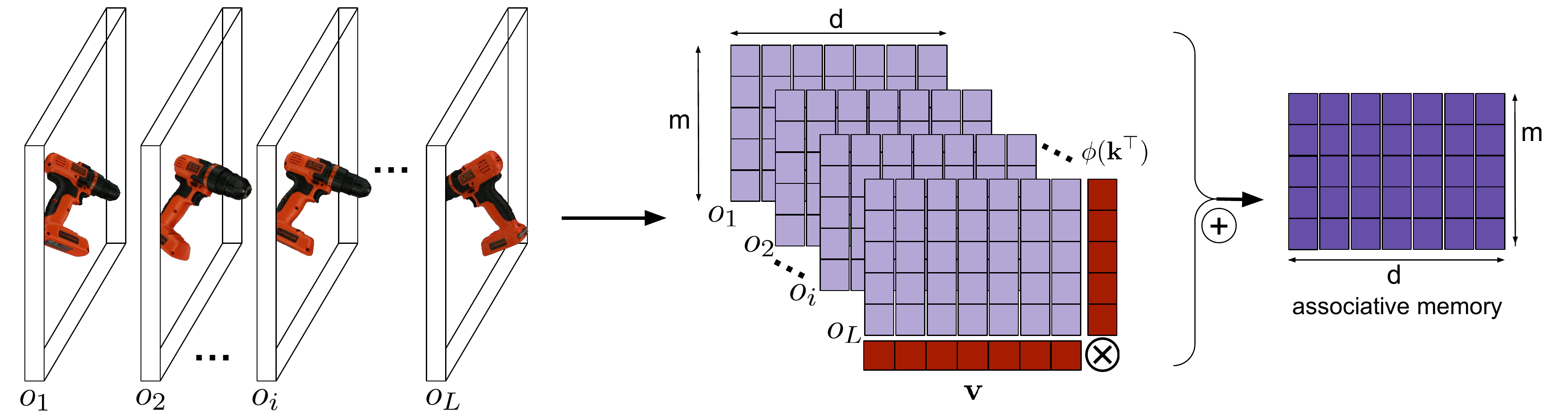}
        \caption{Visual representation of the MVP memory model. The memory (modulo attention-normalization, see Eq.~\ref{eq:compact_memory}) is given by the prefix-sum of the outer-products ($\otimes$) of the $\phi$-transformations of the key-vectors and value-vectors. This memory can be easily accessed and updated as new observations come. The memory effectively compresses all the recorded observations into $o_{1},...,o_{L}$ into bounded space (with size independent of the number of previous recorded observations $L$). The associative memory is updated with each new observation, without requiring a fixed window of previous observations to be tracked. }
        \label{fig:performer_layer_algorithm}
    }
\end{figure} 

To address this, Performers' attention leverages unbiased estimation of the attention-kernel $\mathrm{K}$ via linearization: $\mathrm{K}(\mathbf{q}_{i},\mathbf{k}_{j})=\mathbb{E}[\phi(\mathbf{q}_{i})\phi(\mathbf{k}_{j})^{\top}]$ for some, usually randomized, mapping: $\phi:\mathbb{R}^{d_{\mathrm{QK}}} \rightarrow \mathbb{R}^{m}$. Such a mapping exists for the softmax-kernel in particular (see: FAVOR+/++ mechanisms in \cite{performer,crts}). The linearization leads to the following formula for the approximation of $\mathbf{x}_{i}$:
\begin{align}
\label{eq:compact_memory}
\widehat{\mathbf{x}}_{i} &= 
\sum_{j=1}^{i} \frac{\phi(\mathbf{q}_{i})\phi(\mathbf{k}_{j}^{\top})}{\sum_{l=1}^{i}\phi(\mathbf{q}_{i})\phi(\mathbf{k}_{l})^{\top}}\mathbf{v}_{j} \\ &= 
\frac{\phi(\mathbf{q}_{i})\sum_{j=1}^{i}\phi(\mathbf{k}_{j}^{\top})\mathbf{v}_{j}}{\phi(\mathbf{q}_{i})\sum_{l=1}^{i}\phi(\mathbf{k}_{l})^{\top}} = 
\frac{\phi(\mathbf{q}_{i})\mathbf{M}_{i}}{\phi(\mathbf{q}_{i})\mathbf{m}_{i}},
\end{align}
where $\mathbf{M}_{i} \overset{\mathrm{def}}{=} \sum_{j=1}^{i}\phi(\mathbf{k}_{j})^{\top}\mathbf{v}_{j} \in \mathbb{R}^{m \times d}$ and $\mathbf{m}_{i} \overset{\mathrm{def}}{=} \sum_{l=1}^{i}\phi(\mathbf{k}_{l})^{\top} \in \mathbb{R}^{d_{\mathbf{QK}}}$. We call $\widehat{\mathcal{M}_{i}} = (\mathbf{M}_{i},\mathbf{m}_{i})$ the compact associative memory corresponding to observations $(o_{1},...,o_{i})$. Note that the size of this memory is independent from $i$ and the update of $\widehat{\mathcal{M}_{i}}$ to $\widehat{\mathcal{M}_{i+1}}$ can be also done in time independent from $i$ (see Fig. \ref{fig:performer_layer_algorithm} for the pictorial representation).

In MVP-networks, attention modules of the full Transformer-stacks used in the history-dependent encoder-towers compute latent embeddings of the frames according to Eq.~\ref{eq:compact_memory}. We apply two attention-kernels: regular softmax-kernel (with randomized mapping $\phi$ given by FAVOR+ method) as well as the ReLU kernel defined as: $\mathrm{K}(\mathbf{q}_{i},\mathbf{k}_{j})=\mathrm{ReLU}(\mathbf{q}_{i})\mathrm{ReLU}(\mathbf{k}_{j})^{\top}$ for $\mathrm{ReLU}$ applied element-wise (and trivial corresponding deterministic mapping $\phi$). For more details see~\cite{performer}.

\subsection{The MVP Encoder and Decoder}
The non-attention parts of the MVP encoder and decoder layers are inspired by the CNN architectures described by Varley et al.~\cite{varley2017shapecompletion_iros} and Yang et al.~\cite{Yang18}. Details of the proposed MVP architecture are presented in \autoref{fig:performerv5architecture}. The network takes $L$ unregistered views as $40^3$ voxel-grid inputs, each created by voxelizing a point cloud generated from a 2.5D depth image. All intermediate activation-functions are $\mathrm{RELU}$, and the output activation-function is $\mathrm{sigmoid}$ (since its outputs are in the range $[0,1]$).  The intermediate representation is influenced by both the current observation and all prior observations due to the performer. The output of the Decoder is interpreted as a $40^3$ voxel-grid representing which voxels are occupied by the object independent of whether they are visible to the camera. The output voxel grid is in the same reference frame as the most recent observation. 

Many alternative methods utilize a $32^3$ voxel grid resolution~\cite{what3d_cvpr19,dai2017shape,choy20163d,Peng_2022_CVPR}. We view the use of a $40^3$ voxel grid input and output as an improvement over those alternative methods. Additionally, because the input voxel data is aligned and sized with the output of the object voxel grid, the network is able to reconstruct the pose and shape of an unseen object. When evaluating the occurrence of points in each voxel of the grid, we found that on average an occupied voxel only contained $1.8$ points on average. Given that objects can be placed far from a camera for reconstruction purposes, keeping the voxel resolution at $40^3$ gives enough information for reconstruction of objects.

%% file: sections/experiments.tex
\section{Experiments}

\begin{figure}[t]
    \centering
    \includegraphics[width=\columnwidth]{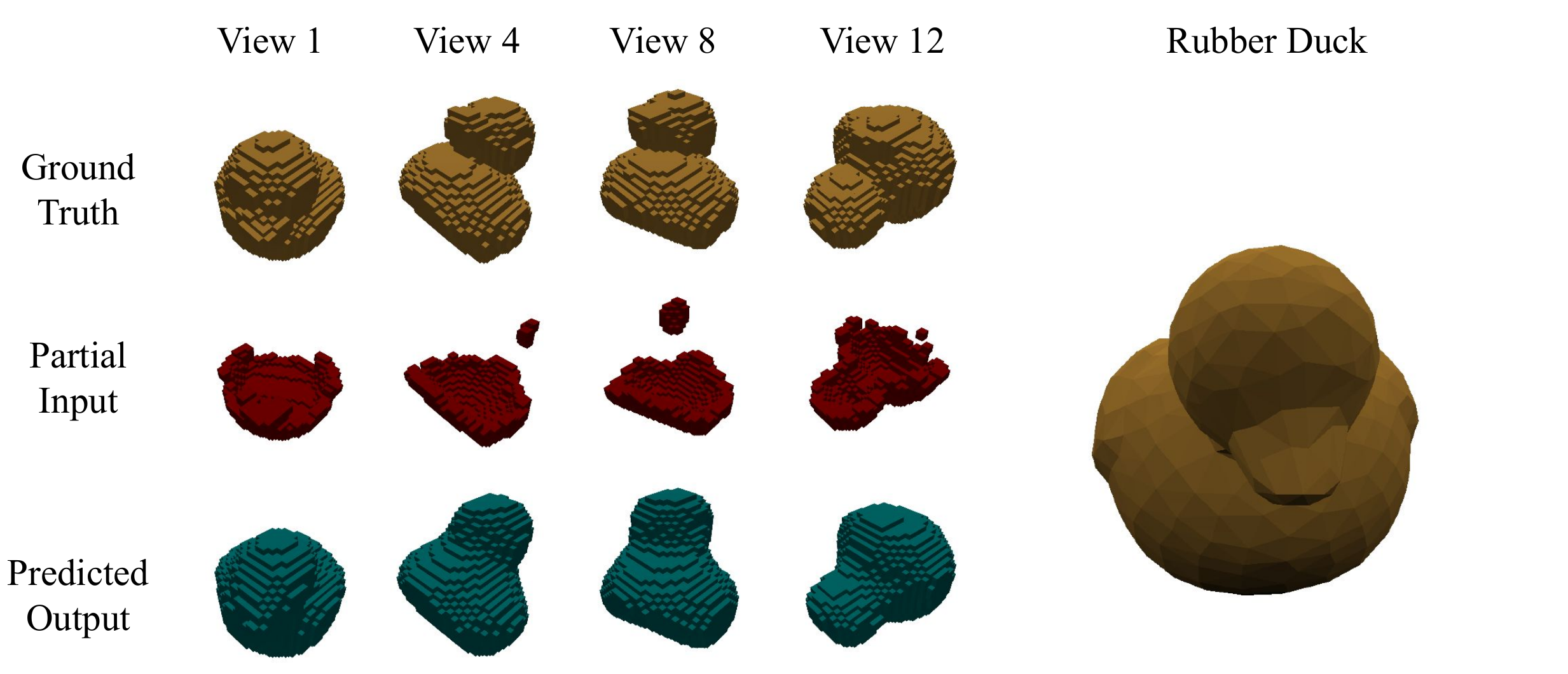}
    \includegraphics[width=\columnwidth]{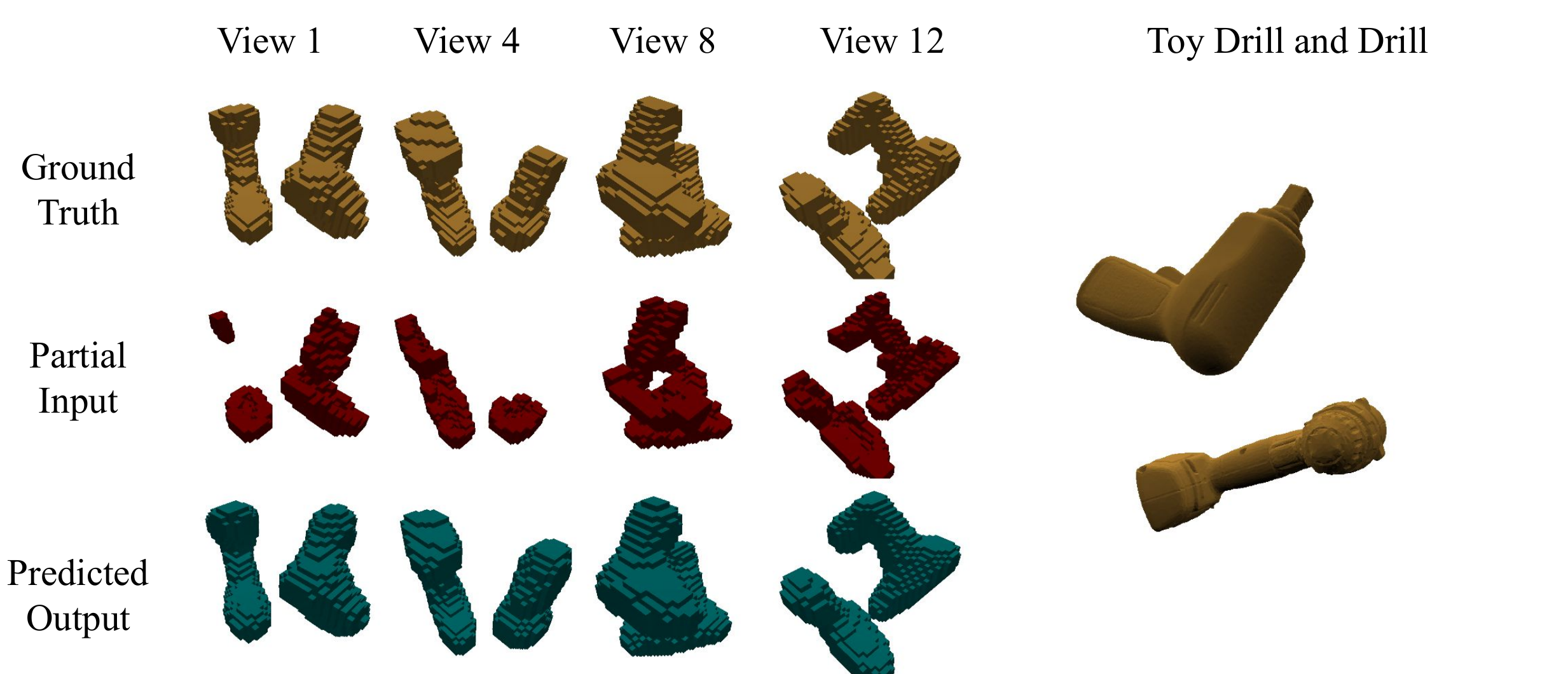}
    \caption{
    (Top) The Camera Pan dataset shows that the MVP model can reconstruct the object at any orientation for each of the $12$ views. (Bottom) The Two Object Camera Pan example shows that the network can robustly complete two objects at the same time. 
    }
    \label{fig:panexamples}
\end{figure}

\begin{figure}[t]
    \centering
    \includegraphics[width=\columnwidth]{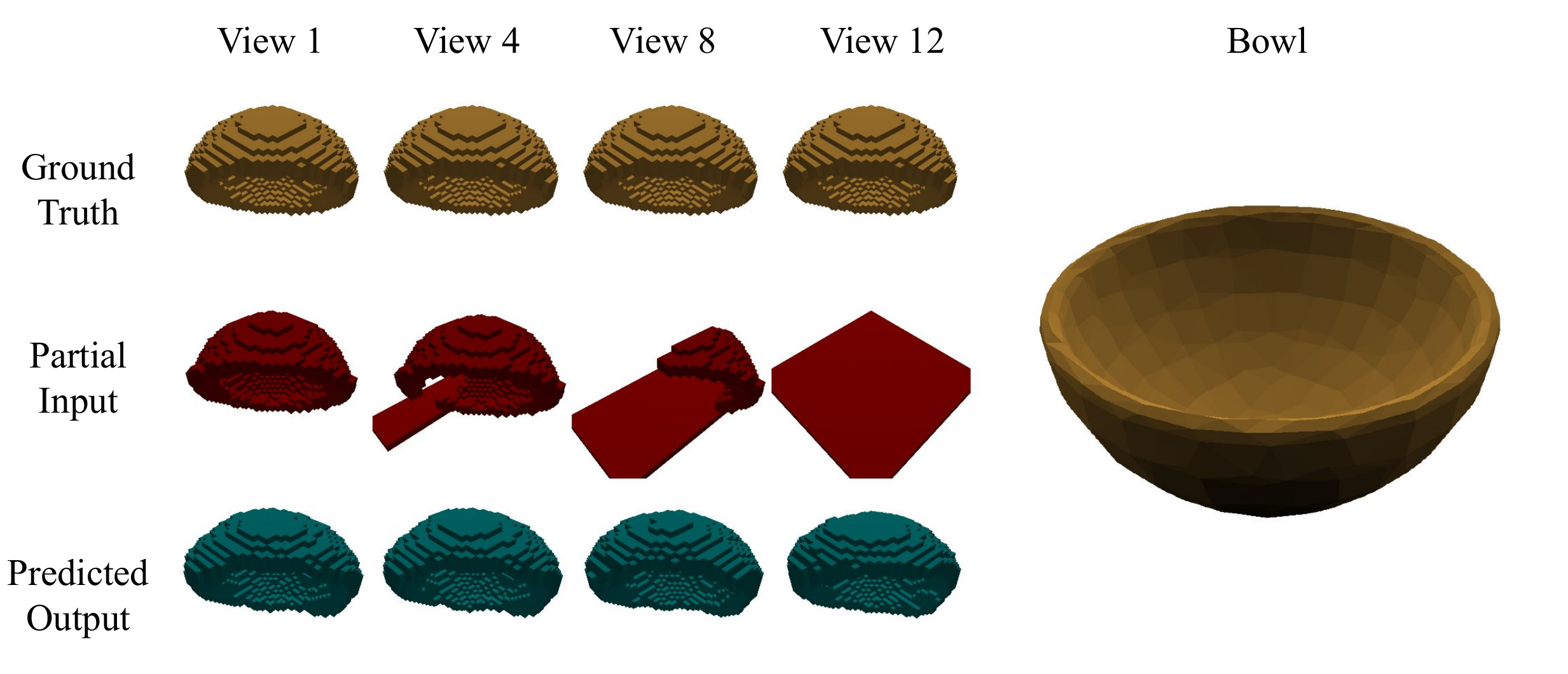}
    \includegraphics[width=\columnwidth]{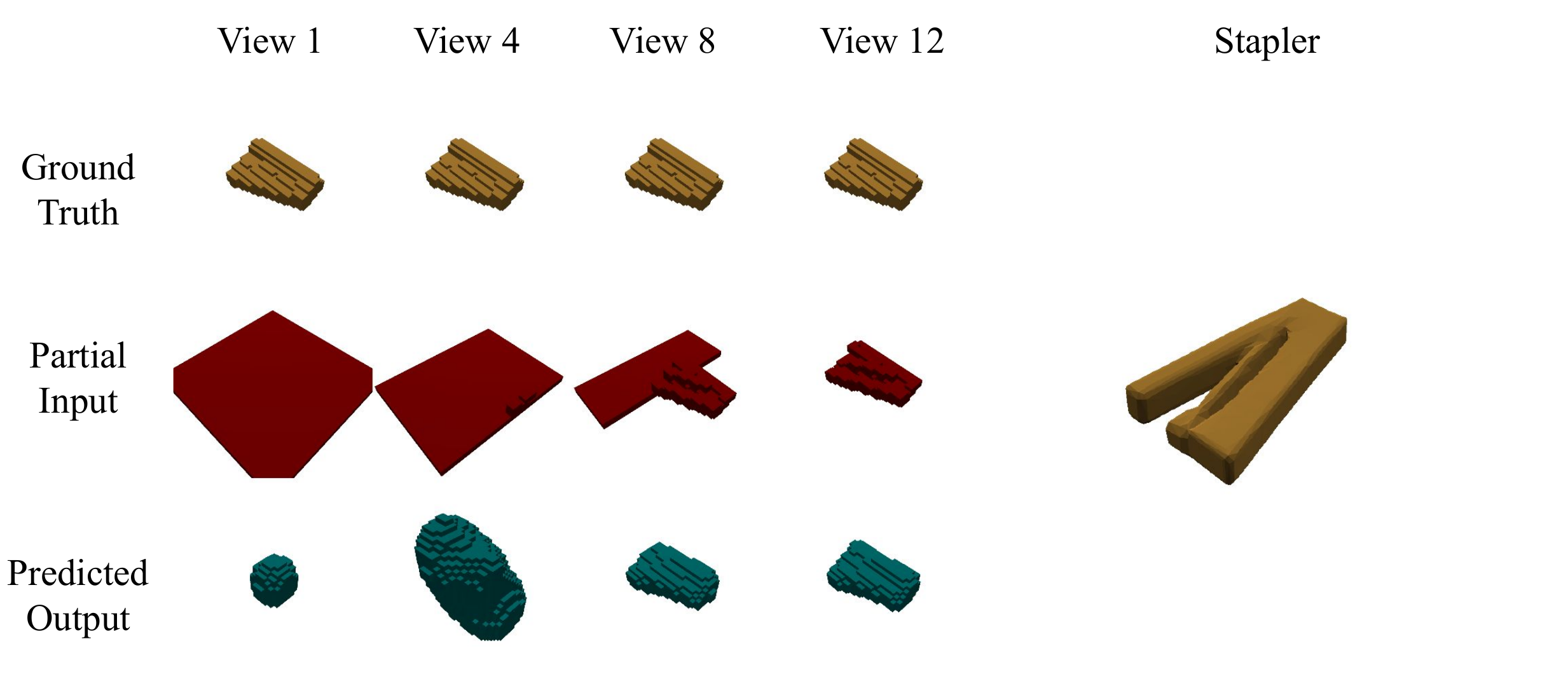}
    \caption{
    (Top) The Object Hiding example shows that the network can remember object geometry despite it no longer being visible with minimal reduction in completion quality. (Bottom) The Object Reveal example shows how the network can incorporate new information about the object as it is slowly revealed.}
    \label{fig:revealhidingexamples}
\end{figure}

\subsection{Baselines}

Three baseline methods are used to evaluate how performers impact the ability to reconstruct objects from multiple views. In the first condition, we evaluate the performance of a single-view reconstruction with a single dense layer in the embedding to evaluate a baseline floor of performance for our method. In the second condition, we modify our MVP architecture to utilize a transformer instead of a performer which we call MVT. In the third condition, we modify our MVP architecture to utilize an LSTM layer instead of a performer. In both of these cases the encoder and decoder are kept constant between Single-View, MVP, LSTM, and MVT conditions. 

Additionally, we evaluate the performance benefit of training with different numbers of views. We evaluate the performance of MVP using 3, 6, and our proposed method of 12 views. We call the MVP model trained with 3 views, MVP3. We call the MVP model trained with 6 views, MVP6.

\subsection{Experimental Setup}

To evaluated performance, twenty-five total models were trained. One for each of the four model architectures (\textit{Single-View}, \textit{MVP}, \textit{LSTM}, \textit{MVT}) for each of the 5 experimental setups). 

\textbf{Camera Pan and Two Object Camera Pan} A camera pans over the object (or pair of objects) capturing a sequence of views. The views are captured by rotating the camera around the centroid of the object.

\textbf{Object Hiding} The object is progressively hidden over time by a sweeping set of voxels that occlude the view. This is evaluating whether the network can continue to output the intended completion after the object has been completely occluded. This experiment is inspired by the concept of object permanence in psychology~\cite{bower1967development}. The object is hidden from view by an incremental $1$ voxel thick curtain and the voxels associated with the object are removed when occluded by the curtain. 

\textbf{Object Reveal} The object is initially hidden from view by a $1$ voxel thick curtain. The object is revealed incrementally as the curtain is removed. The network will demonstrate that it can incorporate the most recent views even if no view has been provided of the object for the first few steps. 

\textbf{Object Slide Behind Other} One object slides behind a stationary object in a the scene becoming partially or fully occluded for several views in the sequence. The start position and distance from the object closest to the camera are varied randomly.

\begin{figure}[]
    \includegraphics[width=\linewidth]{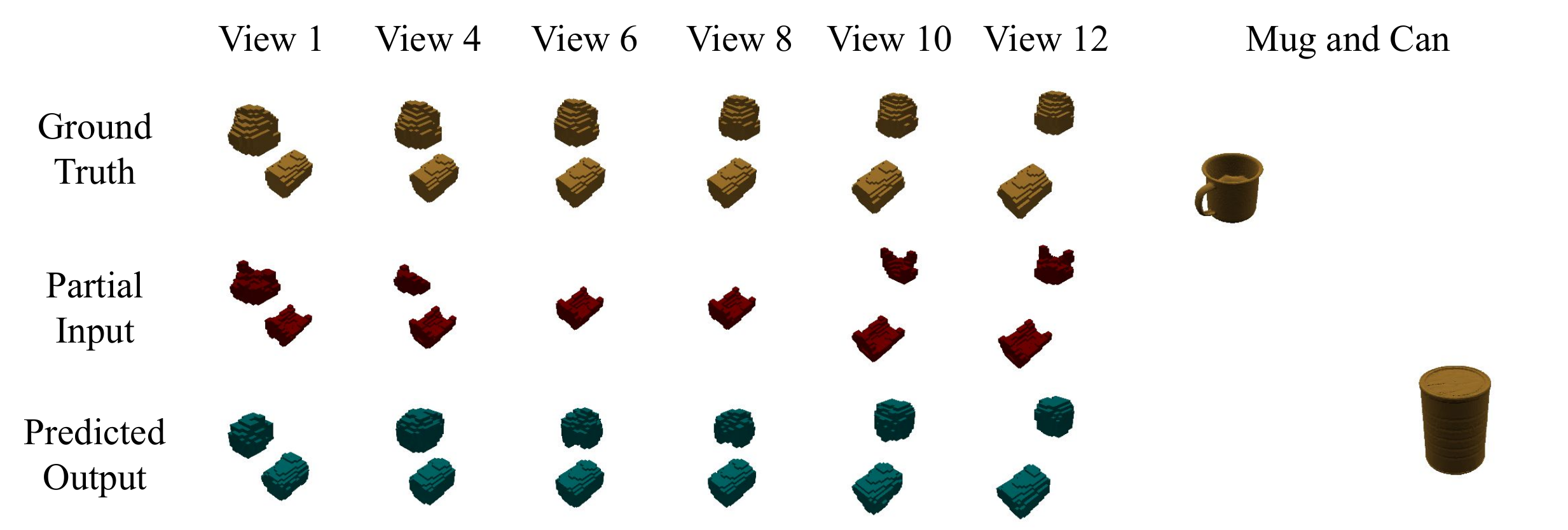}
    \caption{The reconstruction results from the Object Slide Behind Other experimental condition for the MVP model. The Object Slide Behind Other example shows that the MVP network will remember objects even as they are hidden behind an occluding object. All examples shown were not observed during training.}
    \label{fig:sliding_example}
\end{figure}

\begin{figure}[t]
    \centering{
        \includegraphics[width=\linewidth]{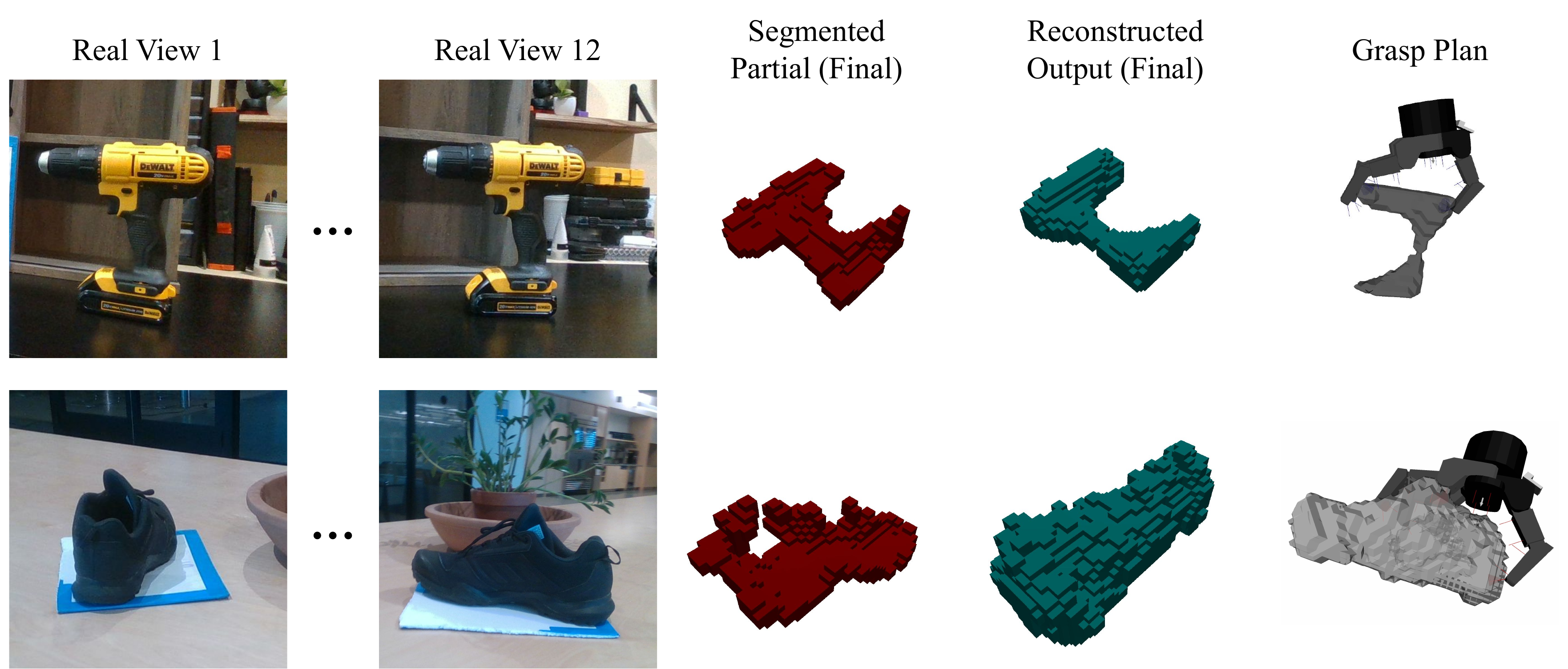}
        \caption{Objects sitting on a table are captured and segmented from their environment. The partial view of each is then reconstructed using the MVP system by capturing multiple views of each object. The reconstructed mesh can then be used in software, such as GraspIt!~\cite{miller2004graspit}, for grasp planning. }
        \label{fig:graspexample}
    }
\end{figure} 

All of these experimental conditions are shown with reconstruction examples in Figures~\ref{fig:panexamples},~\ref{fig:revealhidingexamples}, and~\ref{fig:sliding_example}.

\subsection{Training}

Five different sets of data based on the YCB~\cite{calli2015ycb} and Grasp~\cite{kappler2015leveraging} datasets of objects are used to train the MVP model. Each of the five are based on the five experimental conditions described previously. $55$ objects from the YCB dataset and $463$ objects from the Grasp dataset are used. $726$ views of each object are captured uniformly around each object. For the \textit{Two Object Camera Pan} dataset, a random sample of the $518$ objects are joined as long as they fit in a volume of $0.027m^3$ and then voxelized together. There were $925$ of such pairs from the $518^2$ possible combinations. Each of these pairs are then rendered in simulation about their centroid. For the \textit{Object Slide Behind Other} dataset, $8$ objects were chosen from the YCB dataset and rotated $32$ different orientations for each object, capturing a total of $65536$ sets of $12$ views. Each of the $12$ views capture the object as it slides behind the other. Each of these datasets followed a $80/10/10$ split for train, validation, and test. All views in the test set are of objects not included in the train or validation sets. 

Our MVP model was trained with binary cross-entropy loss and optimized using Adam. For the \textit{Camera Pan} dataset, there were $n=270507$ training samples of $12$ views. 

\subsection{Evaluation}

\begin{table*}[t]
\vspace{5mm}
\centering
\begin{tabular}{|c|rr|rr|rr|rr|rr|}
\hline
                    & \multicolumn{2}{c|}{\textbf{\begin{tabular}[c]{@{}c@{}}Object \\ Hiding\end{tabular}}} & \multicolumn{2}{c|}{\textbf{\begin{tabular}[c]{@{}c@{}}Object \\ Reveal\end{tabular}}} & \multicolumn{2}{c|}{\textbf{\begin{tabular}[c]{@{}c@{}}Camera \\ Pan\end{tabular}}} & \multicolumn{2}{c|}{\textbf{\begin{tabular}[c]{@{}c@{}}Object Slide \\ Behind Other\end{tabular}}} & \multicolumn{2}{c|}{\textbf{\begin{tabular}[c]{@{}c@{}}Two Object \\ Camera Pan\end{tabular}}} \\ \hline
\textbf{Model Name} & \multicolumn{1}{c|}{\textbf{Train}}        & \multicolumn{1}{c|}{\textbf{Test}}        & \multicolumn{1}{c|}{\textbf{Train}}        & \multicolumn{1}{c|}{\textbf{Test}}        & \multicolumn{1}{c|}{\textbf{Train}}       & \multicolumn{1}{c|}{\textbf{Test}}      & \multicolumn{1}{c|}{\textbf{Train}}              & \multicolumn{1}{c|}{\textbf{Test}}              & \multicolumn{1}{c|}{\textbf{Train}}            & \multicolumn{1}{c|}{\textbf{Test}}            \\ \hline
Single-View         & \multicolumn{1}{l|}{N/A}                   & \multicolumn{1}{l|}{N/A}                  & \multicolumn{1}{l|}{N/A}                   & \multicolumn{1}{l|}{N/A}                  & \multicolumn{1}{r|}{0.91}                 & 0.88                                    & \multicolumn{1}{l|}{N/A}                         & \multicolumn{1}{l|}{N/A}                        & \multicolumn{1}{r|}{0.90}                      & 0.87                                          \\ \hline
MVP (ours)          & \multicolumn{1}{r|}{\textbf{0.90}}         & \textbf{0.87}                             & \multicolumn{1}{r|}{\textbf{0.80}}         & \textbf{0.78}                             & \multicolumn{1}{r|}{\textbf{0.96}}        & \textbf{0.90}                           & \multicolumn{1}{r|}{\textbf{0.97}}               & \textbf{0.95}                                   & \multicolumn{1}{r|}{\textbf{0.93}}             & \textbf{0.89}                                 \\ \hline
MVT                 & \multicolumn{1}{r|}{\textbf{0.90}}         & 0.86                                      & \multicolumn{1}{r|}{\textbf{0.80}}         & \textbf{0.78}                             & \multicolumn{1}{r|}{0.95}                 & \textbf{0.90}                           & \multicolumn{1}{r|}{\textbf{0.97}}               & \textbf{0.95}                                   & \multicolumn{1}{r|}{0.92}                      & 0.88                                          \\ \hline
LSTM                & \multicolumn{1}{r|}{\textbf{0.90}}         & 0.86                                      & \multicolumn{1}{r|}{\textbf{0.80}}         & 0.77                                      & \multicolumn{1}{r|}{0.93}                 & \textbf{0.90}                           & \multicolumn{1}{r|}{\textbf{0.97}}               & \textbf{0.95}                                   & \multicolumn{1}{r|}{0.92}                      & 0.88                                          \\ \hline
MVP3                & \multicolumn{1}{r|}{0.89}                  & 0.86                                      & \multicolumn{1}{r|}{0.78}                  & 0.76                                      & \multicolumn{1}{r|}{0.92}                 & \textbf{0.90}                           & \multicolumn{1}{r|}{0.95}                        & 0.94                                            & \multicolumn{1}{r|}{0.91}                      & 0.88                                          \\ \hline
MVP6                & \multicolumn{1}{r|}{0.89}                  & \textbf{0.87}                             & \multicolumn{1}{r|}{0.79}                  & 0.77                                      & \multicolumn{1}{r|}{0.93}                 & \textbf{0.90}                           & \multicolumn{1}{r|}{0.96}                        & 0.95                                            & \multicolumn{1}{r|}{0.92}                      & 0.88                                          \\ \hline
\end{tabular}
\caption{Results for train and test Jaccard. Each Jaccard is the average quality over the $12$ views in each experiment. The MVP model performs at or above the level of the LSTM and MVT models in terms of Jaccard quality. Additionally, the use of 12 views provided benefit over using 3 or 6 views alone. Best results are shown in bold. A higher Jaccard is better. }
\label{tab:completion_table}
\end{table*}

\begin{table*}[]
\centering

\begin{tabular}{|c|cc|cc|cc|cc|cc|}
\hline
                    & \multicolumn{2}{c|}{\textbf{\begin{tabular}[c]{@{}c@{}}Object \\ Hiding\end{tabular}}} & \multicolumn{2}{c|}{\textbf{\begin{tabular}[c]{@{}c@{}}Object \\ Reveal\end{tabular}}} & \multicolumn{2}{c|}{\textbf{\begin{tabular}[c]{@{}c@{}}Camera \\ Pan\end{tabular}}} & \multicolumn{2}{c|}{\textbf{\begin{tabular}[c]{@{}c@{}}Object Slide \\ Behind Other\end{tabular}}} & \multicolumn{2}{c|}{\textbf{\begin{tabular}[c]{@{}c@{}}Two Object \\ Camera Pan\end{tabular}}} \\ \hline
\textbf{Model Name} & \multicolumn{1}{c|}{\textbf{Train}}                   & \textbf{Test}                  & \multicolumn{1}{c|}{\textbf{Train}}                   & \textbf{Test}                  & \multicolumn{1}{c|}{\textbf{Train}}                 & \textbf{Test}                 & \multicolumn{1}{c|}{\textbf{Train}}                         & \textbf{Test}                        & \multicolumn{1}{c|}{\textbf{Train}}                       & \textbf{Test}                      \\ \hline
Single-View         & \multicolumn{1}{c|}{N/A}                              & N/A                            & \multicolumn{1}{c|}{N/A}                              & N/A                            & \multicolumn{1}{c|}{0.90}                           & 0.87                          & \multicolumn{1}{c|}{N/A}                                    & N/A                                  & \multicolumn{1}{c|}{0.9}                                  & 0.87                               \\ \hline
MVP (ours)          & \multicolumn{1}{c|}{\textbf{0.90}}                    & \textbf{0.86}                  & \multicolumn{1}{c|}{\textbf{0.75}}                    & 0.71                  & \multicolumn{1}{c|}{\textbf{0.95}}                  & \textbf{0.90}                 & \multicolumn{1}{c|}{\textbf{0.97}}                          & \textbf{0.95}                        & \multicolumn{1}{c|}{\textbf{0.92}}                        & \textbf{0.88}                      \\ \hline
MVT                 & \multicolumn{1}{c|}{\textbf{0.90}}                    & 0.85                           & \multicolumn{1}{c|}{\textbf{0.75}}                    & 0.71                  & \multicolumn{1}{c|}{0.94}                           & \textbf{0.90}                 & \multicolumn{1}{c|}{\textbf{0.97}}                          & \textbf{0.95}                        & \multicolumn{1}{c|}{0.91}                                 & 0.87                               \\ \hline
LSTM                & \multicolumn{1}{c|}{\textbf{0.90}}                    & 0.85                           & \multicolumn{1}{c|}{\textbf{0.75}}                    & 0.69                           & \multicolumn{1}{c|}{0.92}                           & \textbf{0.90}                 & \multicolumn{1}{c|}{\textbf{0.97}}                          & \textbf{0.95}                        & \multicolumn{1}{c|}{\textbf{0.92}}                        & 0.87                               \\ \hline
MVP3                & \multicolumn{1}{c|}{\textbf{0.89}}                    & \textbf{0.86}                  & \multicolumn{1}{c|}{0.72}                             & 0.70                           & \multicolumn{1}{c|}{0.91}                           & 0.88                          & \multicolumn{1}{c|}{0.95}                                   & 0.94                                 & \multicolumn{1}{c|}{0.91}                                 & 0.87                               \\ \hline
MVP6                & \multicolumn{1}{c|}{0.88}                             & 0.85                           & \multicolumn{1}{c|}{0.74}                             & \textbf{0.72}                  & \multicolumn{1}{c|}{0.92}                           & \textbf{0.90}                 & \multicolumn{1}{c|}{0.96}                                   & \textbf{0.95}                        & \multicolumn{1}{c|}{0.91}                                 & \textbf{0.88}                      \\ \hline
\end{tabular}


\caption{Results for train and test F-score @ 1\%. Each F1-Score is the average quality over the $12$ views in each experiment. The MVP model performs at or above the level of the LSTM and MVT models in terms of F1-Score. Additionally, the use of 12 views provided benefit over using 3 or 6 views alone. Best results are shown in bold. A higher F1-Score is better. }
\label{tab:f1_score_results}
\end{table*}

To evaluate the Single-View, MVP, LSTM, and MVT models, a test dataset of objects not seen during training is reserved for each of the five experimental conditions. Each view is completed and then compared against the ground truth for reconstruction quality. These views are generated by rendering in iGibson~\cite{igibson} and the ground truth is generated by voxelizing the mesh using binvox~\cite{binvox}. Following~\cite{what3d_cvpr19}, we use three metrics to evaluate shape completion quality: Jaccard (MeanIoU)~\cite{jaccard}, F-score @ 1\%~\cite{knapitsch2017tanks}, and grasp joint error~\cite{varley2017shapecompletion_iros}. 

We report the Jaccard similarity metric~\cite{jaccard} as a measure of completion quality. The Jaccard similarity between sets A and B is given by:
\[
J(A, B) = \dfrac{|A\cap B|}{|A\cup B|}
\]
The Jaccard similarity has a minimum value of 0 where A and B have no intersection and a maximum value of 1 where A and B are identical. ~\cite{varley2017shapecompletion_iros} showed that more accurate completions can be helpful for robotic grasp planning.

The F-score is a metric to evaluate the performance of 3D reconstruction results. Our implementation matches~\cite{yang2021single}. It is defined as:
\[
F-Score(d) = \dfrac{2P(d)R(d)}{P(d)+R(d)}
\]
where $P(d)$ and $R(d)$ are the precision and recall with a distance threshold $d$, respectively. The preicison $P(d)$ is defined as:
\[
P(d) = \dfrac{1}{n_\mathscr{R}}\sum_{r\in \mathscr{R}}{}{\min_{g\in\mathscr{G}}||g-r|| < d}
\]

\[
R(d) = \dfrac{1}{n_\mathscr{G}}\sum_{g\in \mathscr{G}}{}{\min_{r\in\mathscr{R}}||g-r|| < d}
\]
where $\mathscr{R}$ and $\mathscr{G}$ represent the predict and ground truth point clouds, respectively. $n_\mathscr{R}$ and $n_\mathscr{G}$ are the number of points in $\mathscr{R}$ and $\mathscr{G}$, respectively. We then convert this voxel grid into a mesh by applying marching cubes to it~\cite{lorensen1987marching} and then extract points from the surface of the mesh. We use this and the ground truth point cloud to calculate a valid F-score. 

We also evaluate the grasp quality of each reconstructed object as described in~\cite{varley2017shapecompletion_iros}. We reconstruct each voxel grid as a mesh using marching cubes~\cite{lorensen1987marching}. We then take that mesh and place it into GraspIt!~\cite{miller2004graspit} and perform an autograsp using a simulated BarrettHand on the mesh. We then place the ground truth mesh in place of the reconstruction and autograsp again. We then calculate the predicted versus realized grasp joint error for each joint and average over each joint. We calculate the grasp joint error for only the final predicted image in each condition. We only use this for completions of single objects, and therefore the \textit{object slide behind other} and \textit{two object camera pan} setups are not evaluated. Examples grasps are shown in~\ref{fig:graspexample}.

\subsection{Results}

Sample reconstructions from MVP tests are shown in Figures~\ref{fig:panexamples},~\ref{fig:revealhidingexamples}, and~\ref{fig:sliding_example}. In all experiments the MVP model can reconstruct images of the object well and remember features of the object that are no longer visible. For the Object Hiding and Object Slide Behind Other experiments, the MVP model was able to remember objects even as they were no longer visible. The enhancement of memory in the neural network architecture of MVP allows for a novel improvement over previous methods. 

The quantitative results for the performer model are shown in Tables~\ref{tab:completion_table},~\ref{tab:f1_score_results}, and~\ref{tab:grasp_joint_error_results}.  The MVP model performs better or equal to the LSTM and MVT models across all experiments in both train and test conditions. The main takeaway is that the MVP model performs similarly to the MVT and LSTM models. In the Object Reveal case, results are lower for the MVP because the first three or four views are empty, which results in a useless completion until information is provided to the network at which point it is able to complete the object well. The most notable improvement is in grasp joint error, where the grasp planner in GraspIt! is very sensitive to object geometry. The improvement in grasp planning shows strong evidence for robotics downstream tasks. These results show MVP's ability to remember objects, generalize to unseen objects, and complete seen objects better than baseline methods. However, all models presented show a significant improvement over a single view. This shows that training using multiple views can improve reconstruction quality in general and is a novel methodology for training shape reconstruction networks. 

%% file: sections/real_world.tex
\section{Real World Completions}

We qualitatively validated that our network can complete objects in a real world as well. To complete objects in the real world a frame is registered to segment the object from nearby surfaces. This landmark is kept track of via an Intel Realsense T265 tracking camera. The object point cloud is captured via an Intel Realsense d415 RGBD camera. This setup is shown in~\autoref{fig:multiviewdemo}. The segmented point cloud is voxelized, then passed through the MVP system to generate an initial object hypothesis. This object hypothesis can then be used in grasp planning software like GraspIt! for robotic grasping, as shown in~\autoref{fig:graspexample}. We found that for the objects we evaluated in the real world that the reconstruction quality was high for the MVP model when compared to the single-view model. 


%% file: sections/conclusion.tex
\section{Conclusion}

\begin{table}[t]
\centering
\begin{tabular}{|c|cc|cc|cc|}
\hline
\multicolumn{1}{|l|}{}                    & \multicolumn{2}{c|}{\textbf{Object Hiding}}                              & \multicolumn{2}{c|}{\textbf{Object Reveal}}                              & \multicolumn{2}{c|}{\textbf{Camera Pan}}                                 \\ \hline
\multicolumn{1}{|l|}{\textbf{Model Name}} & \multicolumn{1}{l|}{\textbf{Train}} & \multicolumn{1}{l|}{\textbf{Test}} & \multicolumn{1}{l|}{\textbf{Train}} & \multicolumn{1}{l|}{\textbf{Test}} & \multicolumn{1}{l|}{\textbf{Train}} & \multicolumn{1}{l|}{\textbf{Test}} \\ \hline
Single-View                               & \multicolumn{1}{c|}{N/A}            & N/A                                & \multicolumn{1}{c|}{N/A}            & N/A                                & \multicolumn{1}{c|}{$4.57^\circ$}           & $4.62^\circ$                               \\ \hline
MVP                                       & \multicolumn{1}{c|}{\textbf{$4.10^\circ$}}  & \textbf{$4.16^\circ$}                      & \multicolumn{1}{c|}{\textbf{$3.80^\circ$}}  & \textbf{$3.94^\circ$}                      & \multicolumn{1}{c|}{\textbf{$3.75^\circ$}}  & \textbf{$3.83^\circ$}                      \\ \hline
MVT                                       & \multicolumn{1}{c|}{$4.14^\circ$}           & $4.19^\circ$                               & \multicolumn{1}{c|}{$3.85^\circ$}           & $3.98^\circ$                               & \multicolumn{1}{c|}{$3.79^\circ$}           & $3.84^\circ$                               \\ \hline
LSTM                                      & \multicolumn{1}{c|}{$4.15^\circ$}           & $4.22^\circ$                               & \multicolumn{1}{c|}{$3.88^\circ$}           & $3.97^\circ$                               & \multicolumn{1}{c|}{$3.82^\circ$}           & $3.89^\circ$                               \\ \hline
MVP3                                      & \multicolumn{1}{c|}{$4.23^\circ$}           & $4.28^\circ$                               & \multicolumn{1}{c|}{$4.00^\circ$}           & $4.09^\circ$                               & \multicolumn{1}{c|}{$3.98^\circ$}           & $4.05^\circ$                               \\ \hline
MVP6                                      & \multicolumn{1}{c|}{$4.18^\circ$}           & $4.22^\circ$                               & \multicolumn{1}{c|}{$3.94^\circ$}           & $3.99^\circ$                               & \multicolumn{1}{c|}{$3.92^\circ$}           & $3.98^\circ$                               \\ \hline
\end{tabular}
\caption{Results for the train and test grasp joint error. Each grasp joint error is the average of all joints in a simulated BarrettHand on the reconstructed object. We find that the performance of the MVP, LSTM, and MVT models are all similar. A lower grasp joint error is better.  }
\label{tab:grasp_joint_error_results}
\end{table}

This paper presented a new approach leveraging multiple unregistered views of an object to predict its mesh with higher accuracy than LSTM and transformer-based models, called \textit{Multiple View Performer} (MVP). MVP critically relies on the scalable implicit-attention Transformers, called Performers, providing compact memory that can be used to utilize previous views of the scene for the shape completion. We demonstrated that MVP is able to remember objects that are no longer visible in the input and to leverage information of the objects that are captured on a delay. We also show that this shape completion system can be used for grasp planning through simulated grasping experiments. All models used in this paper are novel architectures designed to ablate the performance of the MVP. All of these models show that multiple views observed during training result in better performance for shape reconstruction in terms of reconstruction quality and grasp quality metrics. 

%% file: mvp.bbl
\begin{thebibliography}{10}
\providecommand{\url}[1]{#1}
\csname url@samestyle\endcsname
\providecommand{\newblock}{\relax}
\providecommand{\bibinfo}[2]{#2}
\providecommand{\BIBentrySTDinterwordspacing}{\spaceskip=0pt\relax}
\providecommand{\BIBentryALTinterwordstretchfactor}{4}
\providecommand{\BIBentryALTinterwordspacing}{\spaceskip=\fontdimen2\font plus
\BIBentryALTinterwordstretchfactor\fontdimen3\font minus
  \fontdimen4\font\relax}
\providecommand{\BIBforeignlanguage}[2]{{%
\expandafter\ifx\csname l@#1\endcsname\relax
\typeout{** WARNING: IEEEtran.bst: No hyphenation pattern has been}%
\typeout{** loaded for the language `#1'. Using the pattern for}%
\typeout{** the default language instead.}%
\else
\language=\csname l@#1\endcsname
\fi
#2}}
\providecommand{\BIBdecl}{\relax}
\BIBdecl

\bibitem{performer}
\BIBentryALTinterwordspacing
K.~Choromanski, V.~Likhosherstov, D.~Dohan, X.~Song, A.~Gane, T.~Sarl{\'{o}}s,
  P.~Hawkins, J.~Davis, A.~Mohiuddin, L.~Kaiser, D.~Belanger, L.~J. Colwell,
  and A.~Weller, ``Rethinking attention with performers,'' \emph{CoRR}, vol.
  abs/2009.14794, 2020. [Online]. Available:
  \url{https://arxiv.org/abs/2009.14794}
\BIBentrySTDinterwordspacing

\bibitem{shape-completion-base}
\BIBentryALTinterwordspacing
M.~Gualtieri and R.~P. Jr., ``Robotic pick-and-place with uncertain object
  instance segmentation and shape completion,'' \emph{{IEEE} Robotics Autom.
  Lett.}, vol.~6, no.~2, pp. 1753--1760, 2021. [Online]. Available:
  \url{https://doi.org/10.1109/LRA.2021.3060669}
\BIBentrySTDinterwordspacing

\bibitem{shaperetsurvey}
\BIBentryALTinterwordspacing
G.~Fahim, K.~Amin, and S.~Zarif, ``Single-view 3d reconstruction: {A} survey of
  deep learning methods,'' \emph{Comput. Graph.}, vol.~94, pp. 164--190, 2021.
  [Online]. Available: \url{https://doi.org/10.1016/j.cag.2020.12.004}
\BIBentrySTDinterwordspacing

\bibitem{moons}
\BIBentryALTinterwordspacing
T.~Moons, L.~V. Gool, and M.~Vergauwen, ``3d reconstruction from multiple
  images: Part 1 - principles,'' \emph{Found. Trends Comput. Graph. Vis.},
  vol.~4, no.~4, pp. 287--404, 2009. [Online]. Available:
  \url{https://doi.org/10.1561/0600000007}
\BIBentrySTDinterwordspacing

\bibitem{tomo}
F.~Xu and K.~Mueller, ``Real-time 3d computed tomographic reconstruction using
  commodity graphics hardware.'' \emph{Physics in medicine and biology}, vol.
  52 12, pp. 3405--19, 2007.

\bibitem{medicalapp}
\BIBentryALTinterwordspacing
A.~Angelopoulou, A.~Psarrou, J.~G. Rodr{\'{\i}}guez, S.~Orts{-}Escolano, J.~A.
  L{\'{o}}pez, and K.~Revett, ``3d reconstruction of medical images from slices
  automatically landmarked with growing neural models,'' \emph{Neurocomputing},
  vol. 150, pp. 16--25, 2015. [Online]. Available:
  \url{https://doi.org/10.1016/j.neucom.2014.03.078}
\BIBentrySTDinterwordspacing

\bibitem{haefner2019photometric}
B.~Haefner, S.~Peng, A.~Verma, Y.~Qu{\'e}au, and D.~Cremers, ``Photometric
  depth super-resolution,'' \emph{IEEE Transactions on Pattern Analysis and
  Machine Intelligence}, vol.~42, no.~10, pp. 2453--2464, 2019.

\bibitem{varley2017shapecompletion_iros}
J.~Varley, C.~DeChant, A.~Richardson, J.~Ruales, and P.~Allen, ``Shape
  completion enabled robotic grasping,'' in \emph{2017 IEEE/RSJ International
  Conference on Intelligent Robots and Systems (IROS)}, 2017, pp. 2442--2447.

\bibitem{Yang18}
B.~Yang, S.~Rosa, A.~Markham, N.~Trigoni, and H.~Wen, ``Dense 3d object
  reconstruction from a single depth view,'' in \emph{TPAMI}, 2018.

\bibitem{calli2015ycb}
B.~Calli, A.~Singh, A.~Walsman, S.~Srinivasa, P.~Abbeel, and A.~M. Dollar,
  ``The ycb object and model set: Towards common benchmarks for manipulation
  research,'' in \emph{Advanced Robotics (ICAR), 2015 International Conference
  on}.\hskip 1em plus 0.5em minus 0.4em\relax IEEE, 2015, pp. 510--517.

\bibitem{vaswani2017attention}
A.~Vaswani, N.~Shazeer, N.~Parmar, J.~Uszkoreit, L.~Jones, A.~N. Gomez,
  {\L}.~Kaiser, and I.~Polosukhin, ``Attention is all you need,''
  \emph{Advances in neural information processing systems}, vol.~30, 2017.

\bibitem{slim-performers}
\BIBentryALTinterwordspacing
V.~Likhosherstov, K.~M. Choromanski, J.~Q. Davis, X.~Song, and A.~Weller,
  ``Sub-linear memory: How to make performers slim,'' in \emph{Advances in
  Neural Information Processing Systems 34: Annual Conference on Neural
  Information Processing Systems 2021, NeurIPS 2021, December 6-14, 2021,
  virtual}, M.~Ranzato, A.~Beygelzimer, Y.~N. Dauphin, P.~Liang, and J.~W.
  Vaughan, Eds., 2021, pp. 6707--6719. [Online]. Available:
  \url{https://proceedings.neurips.cc/paper/2021/hash/35309226eb45ec366ca86a4329a2b7c3-Abstract.html}
\BIBentrySTDinterwordspacing

\bibitem{hopfield}
\BIBentryALTinterwordspacing
H.~Ramsauer, B.~Sch{\"{a}}fl, J.~Lehner, P.~Seidl, M.~Widrich, L.~Gruber,
  M.~Holzleitner, M.~Pavlovic, G.~K. Sandve, V.~Greiff, D.~P. Kreil, M.~Kopp,
  G.~Klambauer, J.~Brandstetter, and S.~Hochreiter, ``Hopfield networks is all
  you need,'' \emph{CoRR}, vol. abs/2008.02217, 2020. [Online]. Available:
  \url{https://arxiv.org/abs/2008.02217}
\BIBentrySTDinterwordspacing

\bibitem{dope}
\BIBentryALTinterwordspacing
M.~Jin, J.~Li, and L.~Zhang, ``Dope++: 6d pose estimation algorithm for weakly
  textured objects based on deep neural networks,'' \emph{PLOS ONE}, vol.~17,
  no.~6, pp. 1--21, 06 2022. [Online]. Available:
  \url{https://doi.org/10.1371/journal.pone.0269175}
\BIBentrySTDinterwordspacing

\bibitem{watkinsvalls2022mobile}
D.~Watkins-Valls, P.~K. Allen, H.~Maia, M.~Seshadri, J.~Sanabria, N.~Waytowich,
  and J.~Varley, ``Mobile manipulation leveraging multiple views,'' in
  \emph{IROS}, 2022.

\bibitem{miller2004graspit}
A.~T. Miller and P.~K. Allen, ``Graspit! a versatile simulator for robotic
  grasping,'' \emph{IEEE R\&A Magazine}, vol.~11, no.~4, pp. 110--122, 2004.

\bibitem{Wang_2019_CVPR}
C.~Wang, D.~Xu, Y.~Zhu, R.~Martin-Martin, C.~Lu, L.~Fei-Fei, and S.~Savarese,
  ``Densefusion: 6d object pose estimation by iterative dense fusion,'' in
  \emph{Proceedings of the IEEE/CVF Conference on Computer Vision and Pattern
  Recognition (CVPR)}, June 2019.

\bibitem{newcombe2011kinectfusion}
R.~A. Newcombe, S.~Izadi, O.~Hilliges, D.~Molyneaux, D.~Kim, A.~J. Davison,
  P.~Kohi, J.~Shotton, S.~Hodges, and A.~Fitzgibbon, ``Kinectfusion: Real-time
  dense surface mapping and tracking,'' in \emph{Mixed and augmented reality
  (ISMAR), 2011 10th IEEE international symposium on}.\hskip 1em plus 0.5em
  minus 0.4em\relax IEEE, 2011, pp. 127--136.

\bibitem{thrun2008simultaneous}
S.~Thrun and J.~J. Leonard, ``Simultaneous localization and mapping,'' in
  \emph{Springer handbook of robotics}.\hskip 1em plus 0.5em minus 0.4em\relax
  Springer, 2008, pp. 871--889.

\bibitem{krainin2010manipulator}
\BIBentryALTinterwordspacing
M.~Krainin, P.~Henry, X.~Ren, and D.~Fox, ``Manipulator and object tracking for
  in-hand 3d object modeling,'' \emph{The International Journal of Robotics
  Research}, vol.~30, no.~11, pp. 1311--1327, 2011. [Online]. Available:
  \url{https://doi.org/10.1177/0278364911403178}
\BIBentrySTDinterwordspacing

\bibitem{krainin2011autonomous}
M.~Krainin, B.~Curless, and D.~Fox, ``Autonomous generation of complete 3d
  object models using next best view manipulation planning,'' in \emph{2011
  IEEE International Conference on Robotics and Automation}.\hskip 1em plus
  0.5em minus 0.4em\relax IEEE, 2011, pp. 5031--5037.

\bibitem{labbe2019rtab}
M.~Labb{\'e} and F.~Michaud, ``Rtab-map as an open-source lidar and visual
  simultaneous localization and mapping library for large-scale and long-term
  online operation,'' \emph{Journal of Field Robotics}, vol.~36, no.~2, pp.
  416--446, 2019.

\bibitem{3dcnns}
\BIBentryALTinterwordspacing
S.~Xie, C.~Sun, J.~Huang, Z.~Tu, and K.~Murphy, ``Rethinking spatiotemporal
  feature learning: Speed-accuracy trade-offs in video classification,'' in
  \emph{Computer Vision - {ECCV} 2018 - 15th European Conference, Munich,
  Germany, September 8-14, 2018, Proceedings, Part {XV}}, ser. Lecture Notes in
  Computer Science, V.~Ferrari, M.~Hebert, C.~Sminchisescu, and Y.~Weiss, Eds.,
  vol. 11219.\hskip 1em plus 0.5em minus 0.4em\relax Springer, 2018, pp.
  318--335. [Online]. Available:
  \url{https://doi.org/10.1007/978-3-030-01267-0\_19}
\BIBentrySTDinterwordspacing

\bibitem{semantic3dreconstruction}
Zha, J.~Fu, Wang, G.~Baosu, L.~Yin-Sheng, and C.~Yidong, ``Semantic 3d
  reconstruction for robotic manipulators with an eye-in-hand vision system,''
  \emph{Applied Sciences}, vol.~10, p. 1183, 02 2020.

\bibitem{hermann2016eye}
A.~Hermann, F.~Mauch, S.~Klemm, A.~Roennau, and R.~Dillmann, ``Eye in hand:
  Towards gpu accelerated online grasp planning based on pointclouds from
  in-hand sensor,'' in \emph{2016 IEEE-RAS 16th International Conference on
  Humanoid Robots (Humanoids)}.\hskip 1em plus 0.5em minus 0.4em\relax IEEE,
  2016, pp. 1003--1009.

\bibitem{qi2017pointnet}
C.~R. Qi, H.~Su, K.~Mo, and L.~J. Guibas, ``Pointnet: Deep learning on point
  sets for 3d classification and segmentation,'' in \emph{Proceedings of the
  IEEE conference on computer vision and pattern recognition}, 2017, pp.
  652--660.

\bibitem{qi2017pointnet++}
C.~R. Qi, L.~Yi, H.~Su, and L.~J. Guibas, ``Pointnet++: Deep hierarchical
  feature learning on point sets in a metric space,'' \emph{Advances in neural
  information processing systems}, vol.~30, 2017.

\bibitem{irshad2022centersnap}
\BIBentryALTinterwordspacing
M.~Z. Irshad, T.~Kollar, M.~Laskey, K.~Stone, and Z.~Kira, ``Centersnap:
  Single-shot multi-object 3d shape reconstruction and categorical 6d pose and
  size estimation,'' 2022. [Online]. Available:
  \url{https://arxiv.org/abs/2203.01929}
\BIBentrySTDinterwordspacing

\bibitem{siren}
V.~Sitzmann, J.~Martel, A.~Bergman, D.~Lindell, and G.~Wetzstein, ``Implicit
  neural representations with periodic activation functions,'' \emph{Advances
  in Neural Information Processing Systems}, vol.~33, pp. 7462--7473, 2020.

\bibitem{eslami2018neural}
S.~A. Eslami, D.~Jimenez~Rezende, F.~Besse, F.~Viola, A.~S. Morcos, M.~Garnelo,
  A.~Ruderman, A.~A. Rusu, I.~Danihelka, K.~Gregor \emph{et~al.}, ``Neural
  scene representation and rendering,'' \emph{Science}, vol. 360, no. 6394, pp.
  1204--1210, 2018.

\bibitem{texttomesh2022khalid}
N.~M. Khalid, T.~Xie, E.~Belilovsky, and P.~Tiberiu, ``Clip-mesh: Generating
  textured meshes from text using pretrained image-text models,'' December
  2022.

\bibitem{salvi}
\BIBentryALTinterwordspacing
A.~de~Aguiar~Salvi, N.~Gavenski, E.~H.~P. Pooch, F.~Tasoniero, and R.~C.
  Barros, ``Attention-based 3d object reconstruction from a single image,'' in
  \emph{2020 International Joint Conference on Neural Networks, {IJCNN} 2020,
  Glasgow, United Kingdom, July 19-24, 2020}.\hskip 1em plus 0.5em minus
  0.4em\relax {IEEE}, 2020, pp. 1--8. [Online]. Available:
  \url{https://doi.org/10.1109/IJCNN48605.2020.9206776}
\BIBentrySTDinterwordspacing

\bibitem{boyang}
\BIBentryALTinterwordspacing
B.~Yang, S.~Wang, A.~Markham, and N.~Trigoni, ``Robust attentional aggregation
  of deep feature sets for multi-view 3d reconstruction,'' \emph{Int. J.
  Comput. Vis.}, vol. 128, no.~1, pp. 53--73, 2020. [Online]. Available:
  \url{https://doi.org/10.1007/s11263-019-01217-w}
\BIBentrySTDinterwordspacing

\bibitem{mvt}
\BIBentryALTinterwordspacing
S.~Chen, T.~Yu, and P.~Li, ``{MVT:} multi-view vision transformer for 3d object
  recognition,'' \emph{CoRR}, vol. abs/2110.13083, 2021. [Online]. Available:
  \url{https://arxiv.org/abs/2110.13083}
\BIBentrySTDinterwordspacing

\bibitem{danwang}
D.~Wang, X.~Cui, X.~Chen, Z.~Zou, T.~Shi, S.~Salcudean, Z.~J. Wang, and
  R.~Ward, ``Multi-view 3d reconstruction with transformers,'' in
  \emph{Proceedings of the IEEE/CVF International Conference on Computer
  Vision}, 2021, pp. 5722--5731.

\bibitem{choy}
\BIBentryALTinterwordspacing
C.~B. Choy, D.~Xu, J.~Gwak, K.~Chen, and S.~Savarese, ``3d-r2n2: {A} unified
  approach for single and multi-view 3d object reconstruction,'' in
  \emph{Computer Vision - {ECCV} 2016 - 14th European Conference, Amsterdam,
  The Netherlands, October 11-14, 2016, Proceedings, Part {VIII}}, ser. Lecture
  Notes in Computer Science, B.~Leibe, J.~Matas, N.~Sebe, and M.~Welling, Eds.,
  vol. 9912.\hskip 1em plus 0.5em minus 0.4em\relax Springer, 2016, pp.
  628--644. [Online]. Available:
  \url{https://doi.org/10.1007/978-3-319-46484-8\_38}
\BIBentrySTDinterwordspacing

\bibitem{9022145}
T.~Hu, Z.~Han, A.~Shrivastava, and M.~Zwicker, ``Render4completion:
  Synthesizing multi-view depth maps for 3d shape completion,'' in \emph{2019
  IEEE/CVF International Conference on Computer Vision Workshop (ICCVW)}, 2019,
  pp. 4114--4122.

\bibitem{dai2017shape}
A.~Dai, C.~Ruizhongtai~Qi, and M.~Nie{\ss}ner, ``Shape completion using
  3d-encoder-predictor cnns and shape synthesis,'' in \emph{Proceedings of the
  IEEE conference on computer vision and pattern recognition}, 2017, pp.
  5868--5877.

\bibitem{choy20163d}
C.~B. Choy, D.~Xu, J.~Gwak, K.~Chen, and S.~Savarese, ``3d-r2n2: A unified
  approach for single and multi-view 3d object reconstruction,'' in
  \emph{ECCV}.\hskip 1em plus 0.5em minus 0.4em\relax Springer, 2016, pp.
  628--644.

\bibitem{Peng_2022_CVPR}
K.~Peng, R.~Islam, J.~Quarles, and K.~Desai, ``Tmvnet: Using transformers for
  multi-view voxel-based 3d reconstruction,'' in \emph{Proceedings of the
  IEEE/CVF Conference on Computer Vision and Pattern Recognition (CVPR)
  Workshops}, June 2022, pp. 222--230.

\bibitem{lstm-1}
\BIBentryALTinterwordspacing
S.~Hochreiter and J.~Schmidhuber, ``Long short-term memory,'' \emph{Neural
  Comput.}, vol.~9, no.~8, pp. 1735--1780, 1997. [Online]. Available:
  \url{https://doi.org/10.1162/neco.1997.9.8.1735}
\BIBentrySTDinterwordspacing

\bibitem{lstm-2}
\BIBentryALTinterwordspacing
P.~Hoedt, F.~Kratzert, D.~Klotz, C.~Halmich, M.~Holzleitner, G.~Nearing,
  S.~Hochreiter, and G.~Klambauer, ``{MC-LSTM:} mass-conserving {LSTM},'' in
  \emph{Proceedings of the 38th International Conference on Machine Learning,
  {ICML} 2021, 18-24 July 2021, Virtual Event}, ser. Proceedings of Machine
  Learning Research, M.~Meila and T.~Zhang, Eds., vol. 139.\hskip 1em plus
  0.5em minus 0.4em\relax {PMLR}, 2021, pp. 4275--4286. [Online]. Available:
  \url{http://proceedings.mlr.press/v139/hoedt21a.html}
\BIBentrySTDinterwordspacing

\bibitem{gru}
\BIBentryALTinterwordspacing
K.~Cho, B.~van Merrienboer, {\c{C}}.~G{\"{u}}l{\c{c}}ehre, D.~Bahdanau,
  F.~Bougares, H.~Schwenk, and Y.~Bengio, ``Learning phrase representations
  using {RNN} encoder-decoder for statistical machine translation,'' in
  \emph{Proceedings of the 2014 Conference on Empirical Methods in Natural
  Language Processing, {EMNLP} 2014, October 25-29, 2014, Doha, Qatar, {A}
  meeting of SIGDAT, a Special Interest Group of the {ACL}}, A.~Moschitti,
  B.~Pang, and W.~Daelemans, Eds.\hskip 1em plus 0.5em minus 0.4em\relax {ACL},
  2014, pp. 1724--1734. [Online]. Available:
  \url{https://doi.org/10.3115/v1/d14-1179}
\BIBentrySTDinterwordspacing

\bibitem{gru-1}
\BIBentryALTinterwordspacing
J.~Chung, {\c{C}}.~G{\"{u}}l{\c{c}}ehre, K.~Cho, and Y.~Bengio, ``Gated
  feedback recurrent neural networks,'' in \emph{Proceedings of the 32nd
  International Conference on Machine Learning, {ICML} 2015, Lille, France,
  6-11 July 2015}, ser. {JMLR} Workshop and Conference Proceedings, F.~R. Bach
  and D.~M. Blei, Eds., vol.~37.\hskip 1em plus 0.5em minus 0.4em\relax
  JMLR.org, 2015, pp. 2067--2075. [Online]. Available:
  \url{http://proceedings.mlr.press/v37/chung15.html}
\BIBentrySTDinterwordspacing

\bibitem{crts}
\BIBentryALTinterwordspacing
V.~Likhosherstov, K.~Choromanski, A.~Dubey, F.~Liu, T.~Sarlos, and A.~Weller,
  ``Chefs' random tables: Non-trigonometric random features,'' 2022. [Online].
  Available: \url{https://arxiv.org/abs/2205.15317}
\BIBentrySTDinterwordspacing

\bibitem{what3d_cvpr19}
M.~Tatarchenko*, S.~R. Richter*, R.~Ranftl, Z.~Li, V.~Koltun, and T.~Brox,
  ``What do single-view 3d reconstruction networks learn?'' in \emph{CVPR},
  2019.

\bibitem{bower1967development}
T.~Bower, ``The development of object-permanence: Some studies of existence
  constancy,'' \emph{Perception \& Psychophysics}, vol.~2, no.~9, pp. 411--418,
  1967.

\bibitem{kappler2015leveraging}
D.~Kappler, J.~Bohg, and S.~Schaal, ``Leveraging big data for grasp planning,''
  in \emph{ICRA}.\hskip 1em plus 0.5em minus 0.4em\relax IEEE, 2015, pp.
  4304--4311.

\bibitem{igibson}
F.~Xia, W.~B. Shen, C.~Li, P.~Kasimbeg, M.~E. Tchapmi, A.~Toshev,
  R.~Mart{\'\i}n-Mart{\'\i}n, and S.~Savarese, ``Interactive gibson benchmark:
  A benchmark for interactive navigation in cluttered environments,''
  \emph{IEEE Robotics and Automation Letters}, vol.~5, no.~2, pp. 713--720,
  2020.

\bibitem{binvox}
P.~Min, ``binvox,'' {\tt http://www.patrickmin.com/binvox} or \\{\tt
  https://www.google.com/search?q=binvox}, 2004 - 2019, accessed: 2022-05-25.

\bibitem{jaccard}
\BIBentryALTinterwordspacing
S.~Kosub, ``A note on the triangle inequality for the jaccard distance,''
  \emph{Pattern Recognition Letters}, vol. 120, pp. 36--38, 2019. [Online].
  Available:
  \url{https://www.sciencedirect.com/science/article/pii/S0167865518309188}
\BIBentrySTDinterwordspacing

\bibitem{knapitsch2017tanks}
A.~Knapitsch, J.~Park, Q.-Y. Zhou, and V.~Koltun, ``Tanks and temples:
  Benchmarking large-scale scene reconstruction,'' \emph{ACM Transactions on
  Graphics (ToG)}, vol.~36, no.~4, pp. 1--13, 2017.

\bibitem{yang2021single}
S.~Yang, M.~Xu, H.~Xie, S.~Perry, and J.~Xia, ``Single-view 3d object
  reconstruction from shape priors in memory,'' in \emph{Proceedings of the
  IEEE/CVF Conference on Computer Vision and Pattern Recognition}, 2021, pp.
  3152--3161.

\bibitem{lorensen1987marching}
W.~E. Lorensen and H.~E. Cline, ``Marching cubes: A high resolution 3d surface
  construction algorithm,'' in \emph{ACM siggraph computer graphics}, vol.~21,
  no.~4.\hskip 1em plus 0.5em minus 0.4em\relax ACM, 1987, pp. 163--169.

\end{thebibliography}
